\documentclass{article}

\usepackage{PRIMEarxiv}

\usepackage[utf8]{inputenc} 
\usepackage[T1]{fontenc}    
\usepackage{hyperref}       
\usepackage{url}            
\usepackage{booktabs}       
\usepackage{amsfonts}       
\usepackage{nicefrac}       
\usepackage{microtype}      
\usepackage{lipsum}
\usepackage{fancyhdr}       
\usepackage{graphicx}       

\usepackage[title]{appendix}
\usepackage{amsmath}
\usepackage{amsfonts}
\usepackage{amssymb}
\usepackage{graphicx}
\usepackage{siunitx}
\usepackage{xcolor}
\usepackage{booktabs}
\usepackage{caption}
\usepackage{float}
\usepackage{color,soul}
\usepackage{hyperref}

\newcommand{\ocircleplus}{\mathrel{\ooalign{$\bigcirc$\cr\hidewidth$+$\hidewidth\cr}}}

\usepackage{bm}

\usepackage{comment}

\title{Rethinking materials simulations: Blending direct numerical simulations with neural operators} 

\author{
  Vivek Oommen$^1$, Khemraj Shukla$^2$, George Em Karniadakis$^{1,2}$\thanks{corresponding authors}  \\
  1-School of Engineering \\
  2-Division of Applied Mathematics \\
  Brown University \\
  Providence, RI 02912\\
  \texttt{\{vivek\_oommen, khemraj\_shukla, george\_karniadakis\}@brown.edu} \\
   \And
  Saaketh Desai, R\'emi Dingreville\footnotemark[\value{footnote}] \\
  Center for Integrated Nanotechnologies\\
  Sandia National Laboratories\\
  Albuquerque, NM 87185 \\
  \texttt{\{saadesa, rdingre\}@sandia.gov} \\
}

\begin{document}
\maketitle

\begin{abstract}
Direct numerical simulations (DNS) are accurate but computationally expensive for predicting materials evolution across timescales, due to the complexity of the underlying evolution equations, the nature of multiscale spatio-temporal interactions, and the need to reach long-time integration.
We develop a new method that blends numerical solvers with neural operators to accelerate such simulations.
This methodology is based on the integration of a community numerical solver with a U-Net neural operator, enhanced by a temporal-conditioning mechanism that enables accurate extrapolation and efficient time-to-solution predictions of the dynamics.
We demonstrate the effectiveness of this framework on simulations of microstructure evolution during physical vapor deposition modeled via the phase-field method.
Such simulations exhibit high spatial gradients due to the co-evolution of different material phases with simultaneous slow and fast materials dynamics.
We establish accurate extrapolation of the coupled solver with up to 16.5$\times$ speed-up compared to DNS. 
This methodology is generalizable to a broad range of evolutionary models, from solid mechanics, to fluid dynamics, geophysics, climate, and more.

\end{abstract}


\section{Introduction}
\label{sec: intro}
%
{
Materials simulations are omnipresent across diverse scientific domains including physical, chemical, biological, and materials sciences.
Existing state-of-the-art direct numerical solvers used for these simulations, whether they are based on the finite-element \cite{hughes2012finite}, finite-difference \cite{godunov1959finite}, finite-volume \cite{eymard2000finite}, or spectral methods \cite{karniadakis2005spectral}, are inherently computationally expensive because they solve a system of coupled, often nonlinear, partial differential equations (PDEs), requiring a large number of degrees of freedom and advanced numerical schemes to obtain spatially and temporally accurate solutions.
}

{
As an emerging alternative, neural networks are rapidly gaining research interest in the field of scientific computing, especially for learning the solutions of PDEs because of their universal function approximation capabilities~\cite{hornik1990universal}.
Amongst them, physics-informed neural network (PINN) \cite{raissi2019physics} is a specialized and efficient approach for approximating PDEs for forward and inverse problems \cite{hu2023ai, shukla2021physics, kapoor2022predicting, kiyani2023framework, karniadakis2021physics}. 
PINNs incorporate the governing physical laws as soft constraints while training the network parameters, enabling them to act as accurate surrogates in sparse and high-dimensional regimes as well as inverse modeling tasks. 
However, training a physics-informed neural network is still computationally more expensive compared to a data-driven neural network because of the additional computations involved in calculating the PDE constraints from the derivatives of the network outputs with respect to the inputs. 
Therefore, neural networks trained only on data are still used for certain forward problems, depending on the availability of experimental or simulation data. 
Nonetheless, such networks may overfit the available dataset and suffer from high generalization error, which is the error encountered when the trained surrogate network is tested on unseen inputs. 
}

{
From a data-driven learning context, prior works \cite{lu2021learning, li2020fourier} have demonstrated that operator networks trained to approximate mathematical operators can generalize better. 
These operator networks have been developed based on the universal operator approximation theorem by Chen \& Chen \cite{chen1995universal}. 
This theorem extends the approximation capability of neural networks beyond functions to non-linear unbounded operators, which are mappings between infinite-dimensional spaces of functions. 
Deep Operator Networks (DeepONet) \cite{lu2021learning},
Fourier Neural Operators (FNO) \cite{li2020fourier},
Wavelet Neural Operator (WNO) \cite{tripura2023wavelet}, and
Laplace Neural Operator (LNO) \cite{cao2023lno} are some examples of operator networks.
Several research teams \cite{deng2021convergence, lee2023training, de2022generic} have investigated the training procedure, convergence, and generalization errors related to such operator networks. 
Other studies \cite{yin2022simulating, cai2021deepm, mao2021deepm, lin2021operator, wang2021learning, you2022learning, shukla2023deep, kurth2023fourcastnet, goswami2022physics, bora2023learning} have demonstrated the potential of neural operators to serve as accurate surrogates for a broad spectrum of scientific disciplines. 
Nevertheless, surrogate operator networks can be improved further for time-dependent PDEs.
}

{
For example, consider the task of simulating a time-dependent material process by solving a system of non-linear equations over a large computational domain. 
Although direct numerical solvers are the state-of-the-art in terms of accuracy, several applications, such as simulating long time scales or data hungry optimization problems, demand that direct numerical solvers need to be accelerated using faster neural operators that can accurately approximate the underlying mathematical operators over time.
Achieving such acceleration can present a significant challenge since the surrogate operator needs to identify and learn the inherent patterns of the mathematical operator across different spatiotemporal scales from the training dataset and simultaneously exhibit strong generalization capabilities when exposed to new and previously unseen initial conditions. 
U-Net \cite{ronneberger2015u}, a U-shaped fully convolutional neural network, is an interesting choice for modeling such multi-scale systems because of its inherent ability to discretely extract and learn correlations across different scales. 
U-shaped neural operators (U-no) \cite{rahman2022u}, U-FNO \cite{wen2022u} and Diffusion inspired Temporal Transformer Operator (DiTTO) \cite{ovadia2023ditto} are some of the recently developed surrogate operators based on U-Net for non-linear time-dependent PDEs. 
For instance, Gupta \emph{et al.}~\cite{gupta2022towards} extended the concept of positional encoding originally introduced for transformers~\cite{vaswani2017attention} to U-Net-based architectures and performed a detailed systematic ablation study to analyze the significance of Fourier layers \cite{li2020fourier}, attention blocks \cite{vaswani2017attention}, and parameter conditioning \cite{gupta2022towards}. 
The solutions of all the two-dimensional test cases considered in these studies were smooth functions in space. 
The question of whether U-Net-based architectures can be effective in approximating PDEs whose solutions involve high spatiotemporal gradients remains, however, open for research.
}

{
Problems with high spatiotemporal gradients that evolve in time are common place in studies of phase transitions and microstructure evolution of materials~\cite{chen2002phase, stewart2020microstructure, monti2022stability, zhao2023understanding, monismith2023electrochemically, park2023prediction}.
Many research groups~\cite{herman2020data, kibrete2023artificial, kapoor2022surrogate, ryan2022mechanical, shukla2021physics, farizhandi2023spatiotemporal, wu2023emulating, kiyani2023framework, aquistapace2023multisom, he2023novel, alhada2023machine, shin2023deep, kianiharchegani2023data, pagan2022graph} have investigated the effectiveness of using machine-learning-based algorithms for modeling microstructures, with a subset of researchers  
\cite{montes2021accelerating, hu2022accelerating, oommen2022learning, mavi2023unsupervised, zanardi2022adaptive, regazzoni2023latent, desai2024am} proposing to learn the dynamics of such systems in latent spaces.
In this manuscript, we build on these recent advances and develop a method that blends direct numerical solvers with neural operators to accelerate materials simulations.
Specifically, this method is based on the integration of a commonly used phase-field solver~\cite{dingreville2020benchmark, stewart2020microstructure} with a U-Net neural network operator augmented with a temporal conditioning mechanism.
The idea behind using the U-Net operator is to have the ability to learn dynamics across multiple length scales.
The incorporation of the time conditioning is meant to accurately capture the characteristic time scales associated with the various length scales and dynamics across those scales.
We compare this method with two other types of operator networks: a previously proposed autoencoder combined with DeepONet~\cite{oommen2022learning} and a hierarchical autoencoder-based model, which is meant to investigate if hierarchical splitting of the latent space using the proper orthogonal decomposition (POD) modes can further improve the accuracy of the autoencoder-based surrogate operator.
This comparison enables us to conclude that the U-Net with the temporal conditioning mechanism can successfully resolve the multiple spatiotemporal scales better than the other two options.
However, for the long-term evolution of trajectories from new initial conditions, the errors in the predictions provided by the surrogate operator accumulate over time and eventually exceed an acceptable error threshold.
We address this issue by integrating our pre-trained U-Net conditioned operator network with a direct numerical solver to yield an efficient hybrid solver, which accelerates time-to-solution predictions and extrapolates the solution in time.
We demonstrate the comparison of the various operator networks and effectiveness of the hybrid solver on simulations of the microstructure evolution of thin films grown by physical vapor deposition (PVD) modeled via the phase-field method~\cite{dingreville2020benchmark, stewart2020microstructure}.
We chose this problem because it exhibits high spatial gradients that evolve in time due to the co-evolution of materials phases in the growing film and because this problem displays slow (microstructure evolution in the solid film) and fast (moving boundary of the growing film) dynamics simultaneously.  
Furthermore, the vapor deposition processes are used in manufacturing thin films which have widespread engineering applications \cite{cao2009ultrathin,yadav2023morphology}. 
}

{
The manuscript is structured as followed. 
The comparative study of the different surrogate operators,
the implementation of the accelerated hybrid solver, and
its performance
are provided in Section~\ref{sec: results}. 
We discuss
the challenges encountered in data-driven approaches,
the capabilities and limitations of the different surrogate operator networks,
the accuracy-speed trade-off to be considered while designing hybrid solvers, and
future research directions
in Section~\ref{sec: Discussion}. 
The phase-field model,
dataset considered in this study,
description of the training, and
the architecture of the U-Net conditioned on time
are provided in Section~\ref{sec: Materials and Methods}. 
In the appendices, we provide analyses of the simulations by visualizing the evolution of trajectories in the lower dimensional latent space spanned by autoencoder-based models. 
}

\section{Results}
\label{sec: results}
{
We trained our surrogate operator network to learn the spatiotemporal evolution of a materials system that exhibits both slow and fast dynamics. 
Specifically, the operator network is trained to learn the mapping of the composition field $c$ from a set of previous look-back ($lb$) timesteps to $\Delta t$ timesteps in the future in a continuous sense,  
\begin{equation}
    c(t+\Delta t) \approx \mathcal{G}([c(t), ..., c(t-lb+1)])(\Delta t),
    \label{eq: surrogate_formulation}
\end{equation}
where $\mathcal{G}$ represents the operator network trained to approximate the operator that governs the dynamic evolution of $c(t)$. 
Note that $t$ is interpreted as a non-dimensional measure of time throughout the manuscript unless specified explicitly.  
}
\subsection{Comparing and analyzing the different surrogate operator networks}
\label{subsec: comparing_operators}
\begin{figure}
  \centering
  \includegraphics[width=0.75\textwidth]{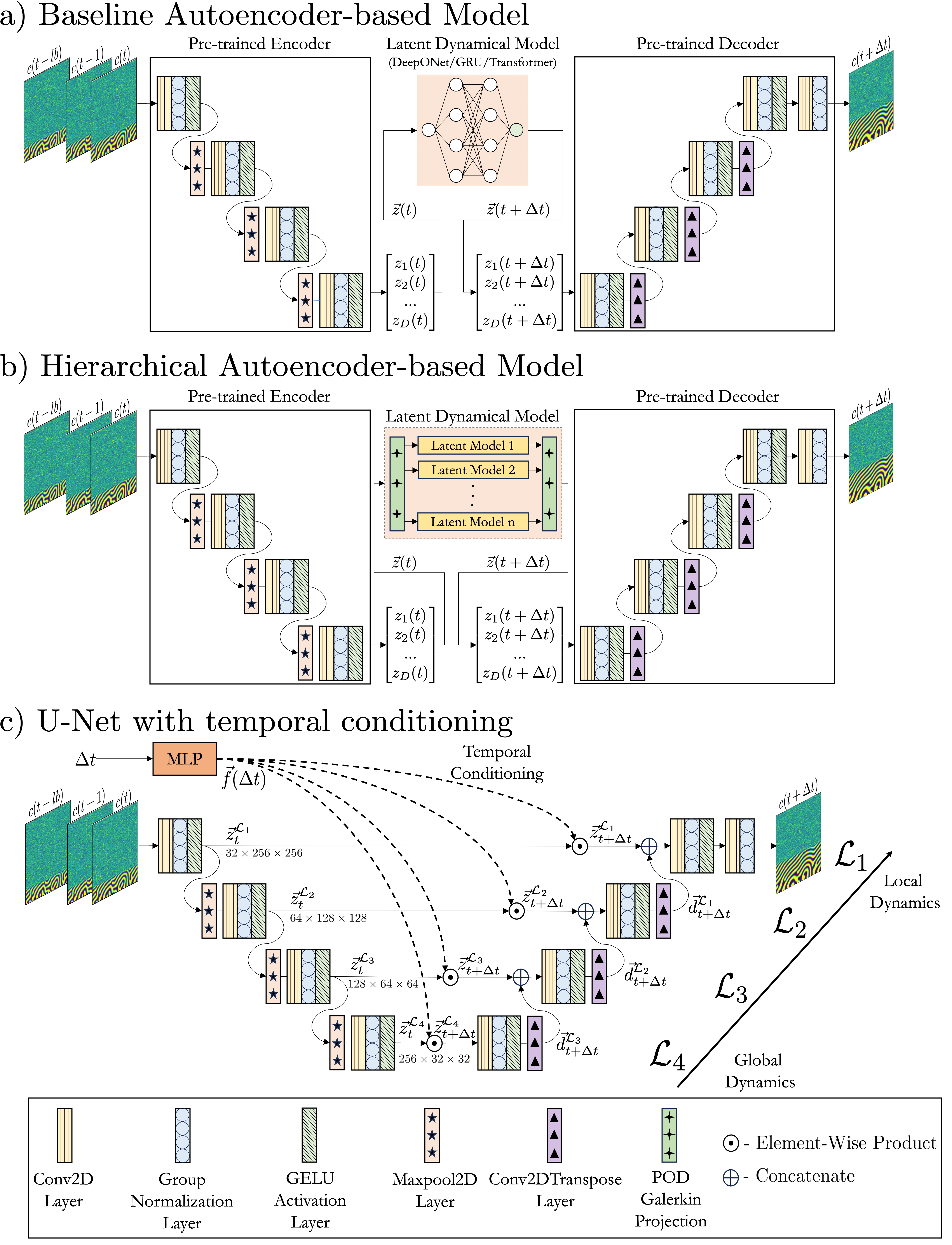}
  \caption{\textbf{Architectures of autoencoder-based models and U-Net with temporal conditioning.} The three types of architectures used in this study that predict the future states, $c(t+\Delta t)$, from a history of previous states, $[c(t-lb+1),...,c(t)]$. a) and b) represent the baseline and hierarchical autoencoder-DeepONet models respectively. In c) a multi-layer perception (MLP) learns a set of basis functions $\vec{f}(\Delta t)$  and conditions the latent representations of $c$ at multiple levels of coarseness ($\mathcal{L}_1$, $\mathcal{L}_2$, $\mathcal{L}_3$ and $\mathcal{L}_4$) inside the U-Net.}
  \label{fig: unet_arch}
\end{figure}

{
In this subsection, we compare the ability of different surrogate operator networks to accurately learn the evolution dynamics of the PVD problem quantitatively.
We consider three operator network architectures: 
1) a previously proposed autoencoder-DeepONet model~\cite{oommen2022learning} which serves as a baseline,
2) a hierarchical autoencoder-based model, and
3) a U-Net with temporal conditioning (see \autoref{fig: unet_arch}). 
In the first two categories of architectures, we first trained an autoencoder to learn a non-linear mapping of the microstructure evolution to a low dimensional latent manifold and then trained another history-dependent surrogate network/networks to learn the evolution dynamics in the low dimensional latent space.
These autoencoder-based models consist of a convolutional encoder and a transpose convolutional decoder paired with one of the following history-dependent neural networks:
Gated Recurrent Unit (GRU) \cite{cho2014learning},
Transformer \cite{giuliari2021transformer}, or
DeepONet \cite{oommen2022learning}
to learn the evolution dynamics in the low-dimensional latent space. 
}

{
The autoencoder-based architectures learn spatial reconstruction first using the encoder and the decoder.
Information on the effect of the size of the dimensionality reduction is provided in Appendix A.
After that, the weights of the encoder and decoder are frozen and the surrogate network is trained to learn the evolution dynamics in the low dimensional latent manifold. 
The autoencoder-based operator networks have one or multiple surrogate dynamical models in a single latent manifold, whereas, the time-conditioned U-Net encodes a representation of time $\vec{f}(\Delta t)$, across multiple latent manifolds -- $\mathcal{L}_1$, $\mathcal{L}_2$, $\mathcal{L}_3$ and $\mathcal{L}_4$, each representing various scales associated with the dynamics of the materials evolution. 
A detailed schematic of the three types of architectures is portrayed in \autoref{fig: unet_arch}.
One may draw an analogy between the V-Grid in the multi-grid method~\cite{hackbusch2013multi} and the temporally conditioned U-Net that learns the underlying global and local dynamics at different resolutions.  
}

\begin{figure}
  \centering
  \includegraphics[width=0.8\textwidth]{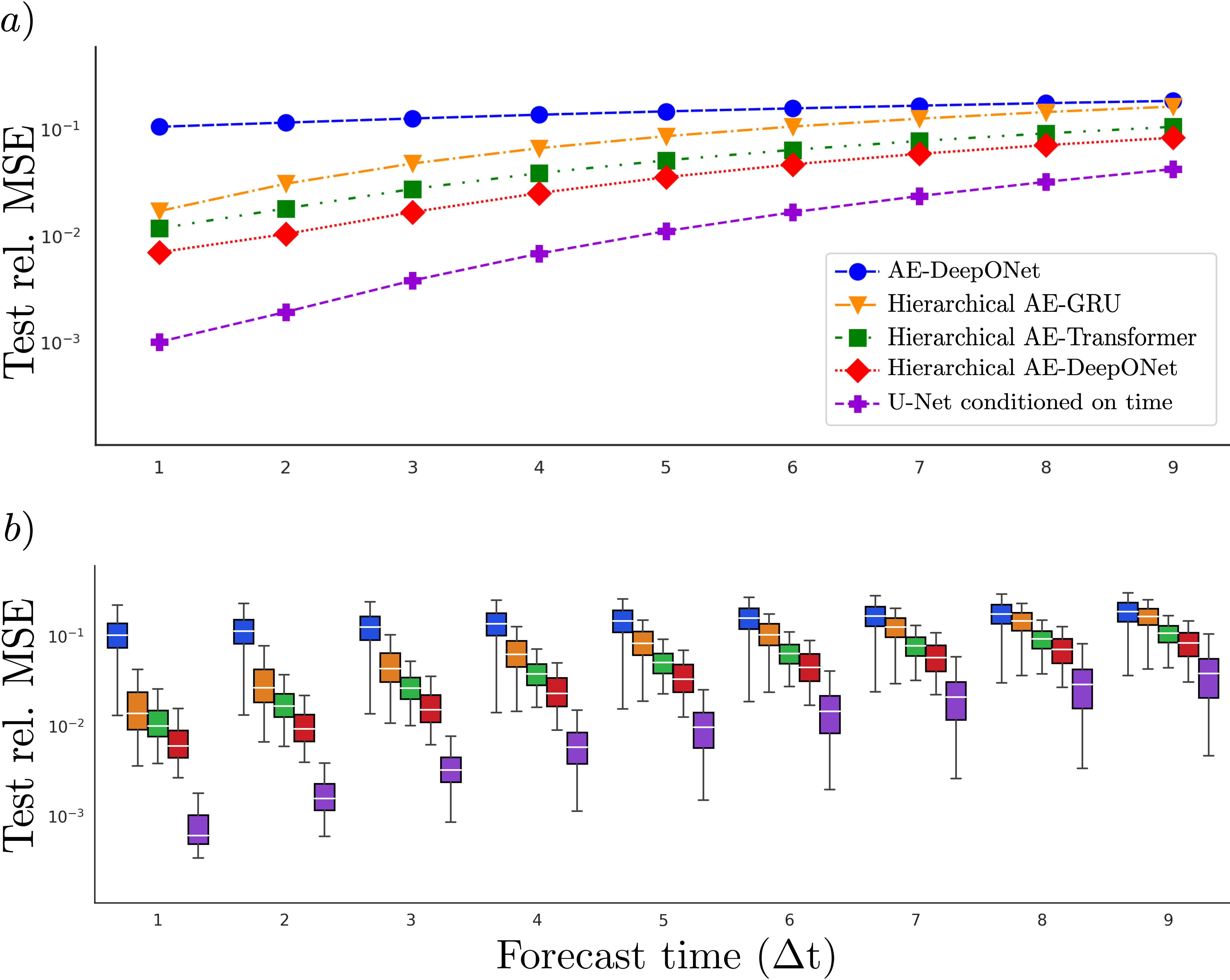}
  \caption{\textbf{Comparing the errors in the forecast.} Relative Mean Square Error (MSE) at different forecast timesteps, computed across all the samples in the test dataset. We compare the prediction errors of autoencoder-based models and U-Net conditioned on time. In a) we observe the general trend in the mean accumulation of test relative MSE, and in b) the error boxplots give us further insights about the distribution of the test relative MSE of each model at each forecast timestep.  }
  \label{fig: error_boxplot}
\end{figure}

{
\autoref{fig: error_boxplot} shows the comparison in prediction performance for the three operator network architectures in terms of the relative mean square error (MSE) (panel a)) and error boxplot (panel b)) at different forecast time $\Delta t$.
Specifically, we considered
1) an autoencoder with a DeepONet as the latent surrogate model~\cite{oommen2022learning},
2) an autoencoder-GRU with hierarchical partitioning of latent space,
3) a hierarchical autoencoder-Transformer,
4) a hierarchical autoencoder-DeepONet, and
5) a temporally conditioned U-Net.
Detailed descriptions of these different models are provided in Appendices B and C. 
All the models were trained using the trajectory-splitting technique described in Appendix D. 
From \autoref{fig: error_boxplot} a) and b), we observe that the hierarchical partitioning of the latent microstructure embedding with respect to the POD modes outperforms the baseline autoencoder-DeepONet model.
Amongst the hierarchical autoencoder-based models, the Transformer predictions are better than that of the GRU because of the self-attention mechanism.
However, both GRU and Transformer models predict the future states autoregressively, leading to an accumulation of errors over time.
The hierarchical DeepONet, on the other hand, is trained to learn the direct mapping into the future states and avoids autoregressive error accumulation.
This attribute enables the DeepONet to predict the future states with lower error than the GRU and transformer that infers autoregressively.
In comparison, the predictions obtained from the U-Net conditioned on time are more accurate than all the autoencoder-based models considered in this study.
This improvement in the performance can be ascribed to the architecture of the network that simultaneously learns the dynamics from low to high-dimensional representations of the system.
Despite these improvements, surrogate operator networks trained in a data-driven setting, although faster at inference, may not necessarily be suitable for forecasting long time scales.
We address this difficulty in the following subsection by blending the phase-field numerical solver (\textbf{Me}soscale \textbf{M}ultiphysics \textbf{Ph}ase F\textbf{i}eld \textbf{S}imulator or MEMPHIS in short, see Refs.~\cite{stewart2020microstructure,dingreville2020benchmark}) with our best surrogate operator - the pre-trained U-Net conditioned on time as a hybrid solver.
}

\subsection{Coupling the high-fidelity direct numerical solver with an operator network-based surrogate}
\label{subsec: hybrid_solver}
\begin{figure}
  \centering
  \includegraphics[width=0.99\textwidth]{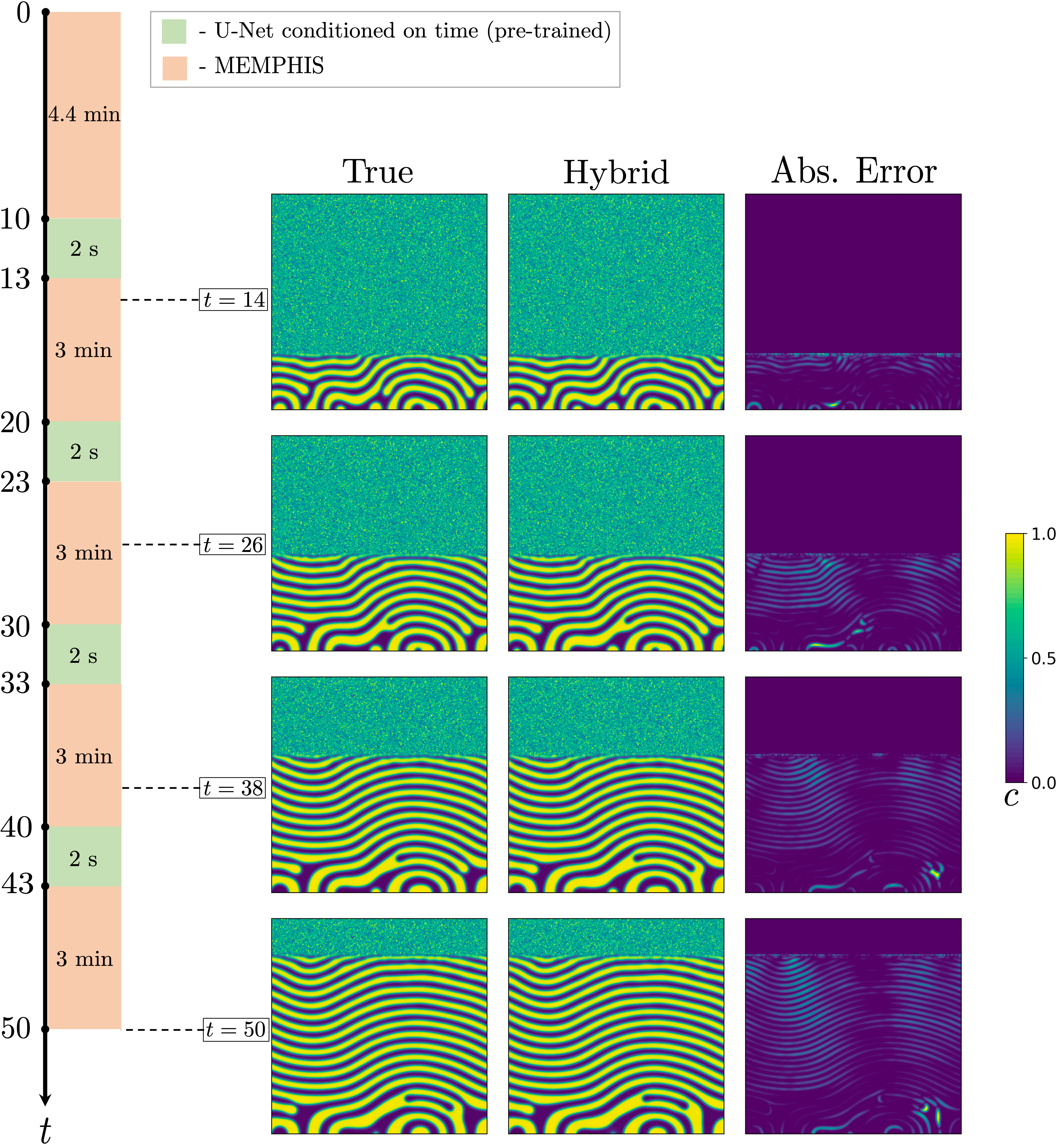}
  \caption{\textbf{Hybrid trajectory with 1.3x speed-up.} The left column of images represents a true trajectory of the normalized composition field $c$ generated by our direct numerical solver MEMPHIS. The middle column shows the trajectory from the same initial condition predicted by the U-Net hybrid model consisting of a pre-trained U-Net conditioned on time. Note that this is a sample from the test dataset, not seen by the surrogate conditioned U-Net during the training stage. The details about the back-and-forth transitions and the computational time on an 8-core CPU can be seen on the time axis placed on the left side of the figure. The right column represents the evolution of the absolute pointwise error. (Check \href{https://github.com/vivek-brown/UNet-with-temporal-conditioning} {git repository} for animation.)}
  \label{fig: hybrid_trajectory}
\end{figure}

{
Using an operator network-based surrogate for forecasting longer in time is a challenge and is extensively investigated in various prior works \cite{montes2021accelerating, oommen2022learning, fernex2021cluster}.
This predicament is especially relevant in the PVD exemplar, where the interface of the growing film evolves rapidly compared to the slower diffusion-driven evolution of species in the bulk of growing film.
Therefore, surrogate operator networks trained within a data-driven framework, despite their fast inference, might not be appropriate for making extended forecasts due to the gradual accumulation of errors over time.
We address this problem of error accumulation over time via a hybrid formulation that integrates the accurate but slower finite-difference-based direct numerical solver with the pre-trained U-Net conditioned on time.
The purpose of this hybrid approach is to accelerate the direct numerical simulations with the surrogate operator network by leaping ahead in time, periodically intervening with the direct numerical solver to keep numerical errors and physical solutions bounded.
Specifically, we run our phase-field solver (MEMPHIS) for the first few timesteps.
The output from the direct numerical simulation is then fed as an input to the pre-trained U-Net conditioned on time.
The fast inference capability of the conditioned U-Net enables the hybrid formulation to leap in time for a prescribed time interval.
The output of the surrogate operator is fed back as input to the phase-field solver, which is again executed for a certain number of timesteps, before leaping forward through the U-Net.
In this manner, the U-Net hybrid model orchestrates the evolution trajectory of the microstructure's composition field through back-and-forth transitions between the phase-field solver and the temporally conditioned U-Net.
For example, in \autoref{fig: hybrid_trajectory}, the orange shade along the time axis $t$, corresponds to the timesteps evolved by the direct numerical phase-field solver, and the green shade corresponds to the timesteps accelerated by the pre-trained surrogate U-Net conditioned on time.
The wall time associated with running the U-Net hybrid model on an 8-core CPU is also provided.
Without trying to optimize the partitioning between time spent on the direct numerical solver and time spent on the operator network-based surrogate, we observe that the hybrid solver achieves a speed-up of 30\%.
Furthermore, from the comparison between the evolution of true and hybrid trajectories in \autoref{fig: hybrid_trajectory}, we observe that the hybrid solver is able to predict the dynamics of the composition field $c$, with good accuracy over time and space for an unseen test sample.
}

\subsection{Analyzing the local prediction errors of the hybrid trajectories}
\label{subsec: autocorrelations}
\begin{figure}
  \centering
  \includegraphics[width=0.9\textwidth]{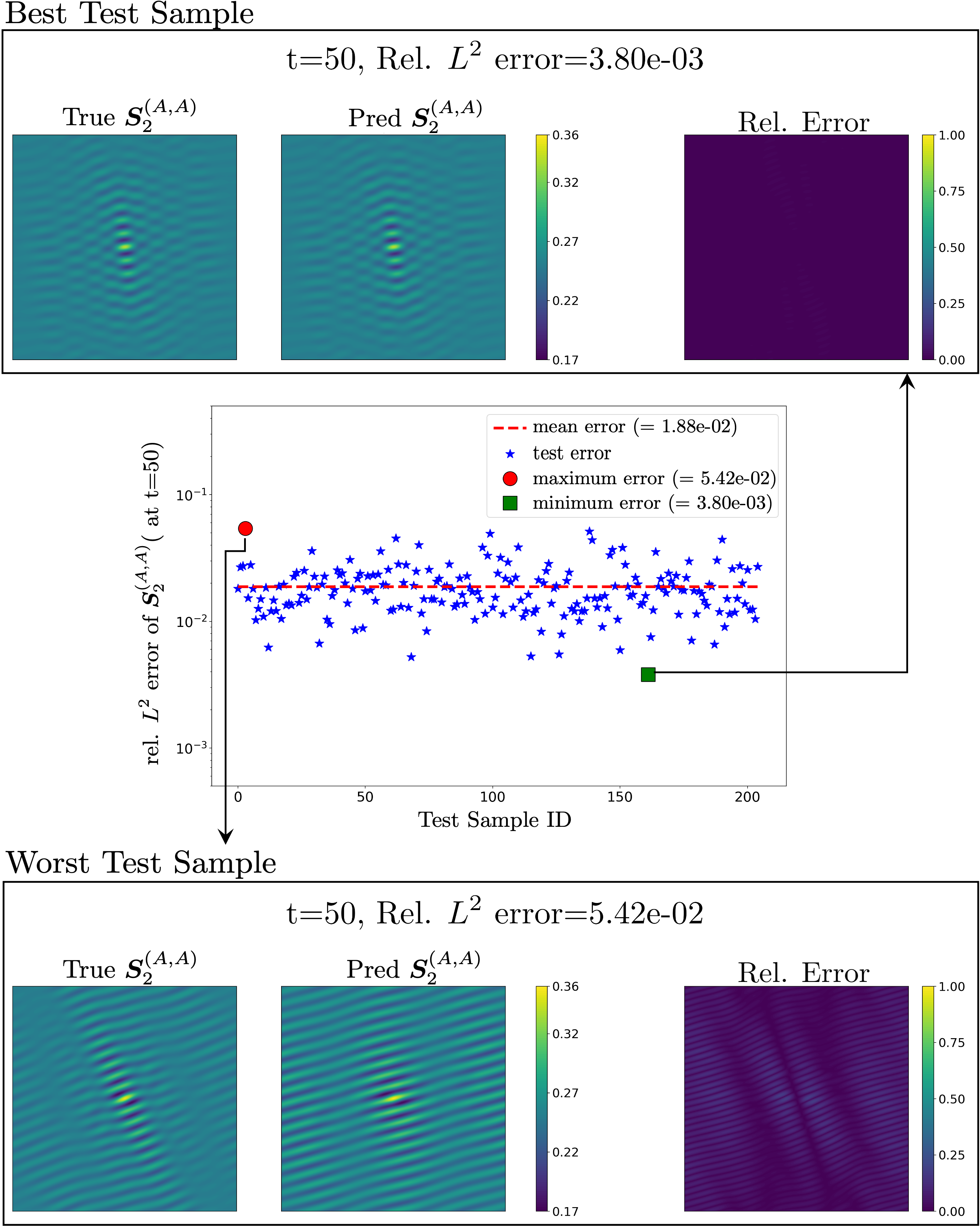}
  \caption{\textbf{Errors in the autocorrelations of the microstructure.} Scatter plot to visualize the relative $L^2-$error of $\bm{S}_2^{(A,A)}(\bm{r},t=50)$ for all the samples from the test dataset. From the scatter plot, we identify the worst test sample with the largest error and the best test sample with the lowest error and plot the true and predicted autocorrelations at $t=50$, along with its absolute point-wise error.}
  \label{fig: error_statistics}
\end{figure}

\begin{figure}
  \centering
  \includegraphics[width=0.99\textwidth]{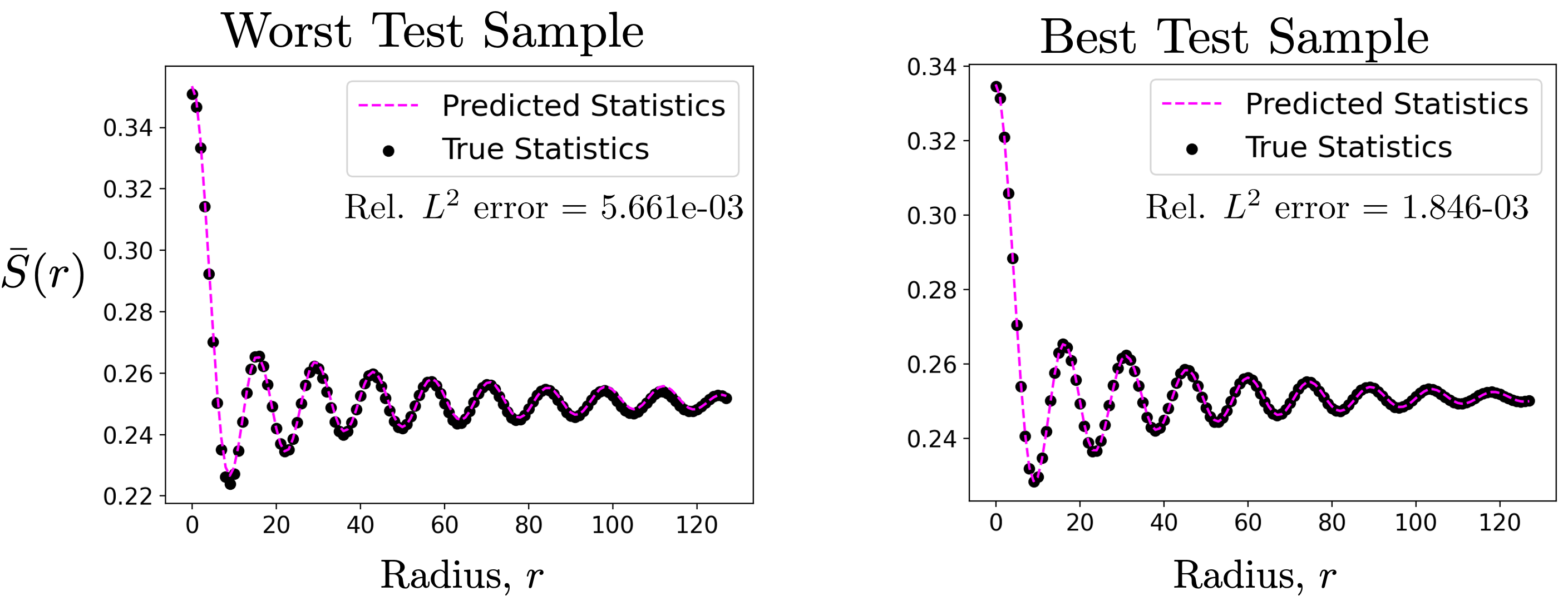}
  \caption{\textbf{Error in the radial average of autocorrelations.} Radial average of the autocorrelations for the best and the worst test samples identified from the scatter plot in Figure~\ref{fig: error_statistics}}. 
  \label{fig: r_avg_error_statistics}
\end{figure}

{
We now evaluate the local prediction errors of the microstructure evolution obtained from the hybrid solver.
By computing the autocorrelations of the composition field on the solid phase of the growing film, we can extract and visualize the prominent features and patterns in the microstructure.
The spatial autocorrelation, $\bm{S}_2^{(A,A)}(\bm{r},t)$, of the composition field with two species $A$ and $B$ can be interpreted as the probability of finding the same solid phase $A$ at two random spatial locations $\bm{x}_1$ and $\bm{x}_2$, where $\bm{r} = \bm{x}_2 - \bm{x}_1$.
In \autoref{fig: error_statistics}, we show the relative $L^2-$error with respect to $\bm{S}_2^{(A,A)}(\bm{r},t=50)$ for all the trajectories at the last timestep in the test dataset, since the maximum error accumulation is present at the final time.
We observe that the mean relative $L^2-$error of autocorrelation of the U-Net hybrid model's composition field at $t=50$ is 0.0188.
Furthermore, we compare and visualize the true and predicted $\bm{S}_2^{(A,A)}(\bm{r},t=50)$, and its point-wise error for the test sample with a maximum error of 0.0542 representing the worst case and that of the test sample with a minimum error of 0.0038.
Next, we compute and compare the radial average of the autocorrelations $\bar{S}(r,t=50)$ in \autoref{fig: r_avg_error_statistics} corresponding to the worst and the best test samples identified from the scatter plot in \autoref{fig: error_statistics}. 
Here again the errors are small, with relative $L^2-$errors of 0.0057 and 0.0018, respectively.
We observe that the predicted statistics capture all the correlation lengths present in the microstructure with great accuracy, even for the `worst' test sample, demonstrating the generalization capability of the hybrid solver.
Overall, we observe a very good agreement between the predicted and true autocorrelations, demonstrating that the hybrid solver is a fast solver with excellent spatial and temporal accuracy across time.
}

\subsection{The need for hybrid solvers}
\label{subsec: need_for_hybrid}
\begin{figure}
  \centering
  \includegraphics[width=0.8\textwidth]{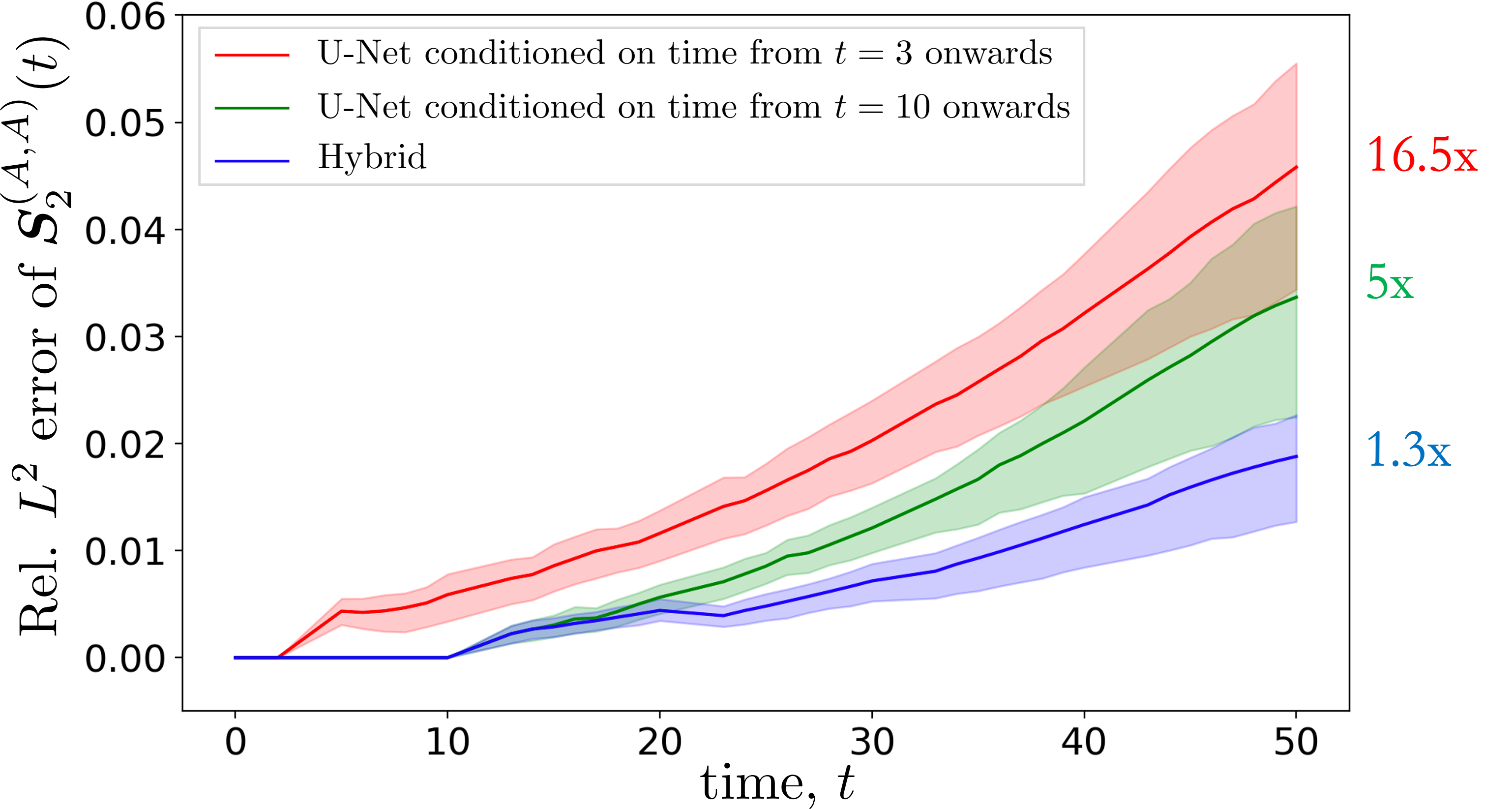}
  \caption{\textbf{Time evolution of autocorrelation error in hybrid trajectories.} The plot provides insights on how the relative $L^2-$error associated with autocorrelation of microstructure predicted by the conditioned U-Net hybrid model accumulates over time. The solid line is the mean error and the shaded regions represent the 25th to 75th error quartiles computed across all the samples in the test dataset. The hybrid formulation involves multiple back-and-forth transitions between the numerical solver and the surrogate as shown in \autoref{fig: hybrid_trajectory}. }
  \label{fig: error_evolution}
\end{figure}

{
In the previous subsections, we investigated the errors in the autocorrelations and the radial average of the autocorrelations of the microstructure predicted by the U-Net hybrid model at the last timestep. 
We now investigate how the error in the autocorrelations of the microstructure predicted by the surrogate U-Net conditioned on time propagates in time.
This is especially important in the context of understanding how quickly the predicted solution starts to deviate from the ground-truth solution and evaluating the trade-off between speed-up and accuracy.
To accomplish this evaluation, we tested several configurations.
For the first set of predictions, we ran our direct numerical phase-field solver until only $t=3$ and then used the trained U-Net conditioned on time from $t=3$ onward.
The second set of predictions considered the direct numerical phase-field solver that ran until $t=10$ and then used U-Net conditioned on time to roll out the remaining part of the trajectory.
Finally, the last set of predictions consisted of the hybrid solver presented in Section~\ref{subsec: hybrid_solver}.
The evolution of mean error as well as the 25th and 75th quartiles of the error with respect to time for these three sets of prediction are plotted in \autoref{fig: error_evolution}.
For all three configurations, we note first that the error accumulates as a function time.
We also observe that the two configurations for which the U-Net conditioned on time replaces the direct numerical solver (red and green curves in \autoref{fig: error_evolution}) have similar rates of error growth as the U-Net marches in time.
This result suggests that running the direct numerical solver for longer timesteps helps in controlling the accumulation of errors in forecasting, albeit reducing the overall speedup achieved.
When contrasting these results with the hybrid solver implementation (in blue in in \autoref{fig: error_evolution}), we observe that the back-and-forth transitions between the solver and the surrogate helps bringing down the error, the rate of accumulation of error in time, and also the variance of error.
In general, the results shown in \autoref{fig: error_evolution} suggest that the accumulation of error in prediction can be reduced by running the solver for longer timesteps, but at the expense of higher computational costs and therefore reduced acceleration performance.
Indeed, in the case of the first configuration for which the U-Net conditioned on time replaces the direct numerical solver early only we achieve the highest speedup with 16.5$\times$ but with the largest error.
The other similar configuration but for which the direct numerical solver runs for longer time before being replaced achieved a speedup of 5$\times$.
Finally, our hybrid solver achieved a 1.3$\times$ speedup, however, we did not attempt to optimize the partitioning between DNS and the operator network. 
}

\section{Discussion}
\label{sec: Discussion}

{
The results presented above demonstrate the effectiveness of using a U-Net with temporal conditioning as a fast and accurate surrogate model for problems with high spatial gradients and simultaneous fast and slow dynamics.
However, training a surrogate neural network model for learning this type of problems is challenging because of two main reasons: 
the high-gradient regions that evolve in time and the ability to accurately capture both the slow and fast dynamics.
In the present case of the PVD simulations, the microstructure evolution in the growing film represents the slow dynamics and the moving interface of the solid thin film growing represents the fast dynamics.
This means that the surrogate network must learn accurately the evolution of the moving solid-vapor interface with time. 
Any error in modeling the interfacial dynamics introduces errors in the predicted microstructure that only gets worse during the remaining course of evolution.
At the same time, because of the duality of dynamics, the surrogate network should also be capable of learning and accurately modeling the multi-scale non-linear dynamics, rendering the training process delicate.
}

{
The first challenge of learning the evolution dynamics involving several high spatial gradient regions can be addressed by training an autoencoder that learns a suitable mapping to a lower dimensional latent space, and then training a surrogate latent dynamical model to learn the evolution dynamics of the composition field in the latent space of the autoencoder \cite{oommen2022learning}.
We extended this approach to learning the latent dynamics with greater accuracy by hierarchically partitioning the latent space with respect to POD modes.
More details regarding implementation and experiments can be found in the Appendix.
However, in our PVD exemplar problem, such autoencoder-based models may not be a suitable choice for modeling the evolution of the solid-vapor interface because of the interface boundary conditions describing that interface.
Indeed, in the PVD problem, the boundary condition at the interface is a random draw from a truncated Gaussian distribution and therefore behaves like a noise.
The encoder part of the autoencoder, comprising the two-dimensional convolutional layers, smooths out the noise-like interface boundary condition resulting in the loss of information.
Therefore, the autoencoder-based models that learn the evolution dynamics in a single latent manifold might not be a good choice for modeling problems with similar challenges.
On the contrary, the U-Net conditioned on time learns the inherent dynamics of the system in multiple latent manifolds, namely $\mathcal{L}_1$ to $\mathcal{L}_4$.
Therefore, the temporally conditioned U-Net is able to take advantage of the information at the solid-vapor interface and overcome the difficulty of over-smoothing the interface boundary condition encountered in the autoencoder-based models.
The results in Section~\ref{sec: results} show that the U-Net conditioned on time can effectively learn both the fast-evolving interfacial dynamics and the slower diffusion-driven dynamics accurately, compared to the autoencoder-based models.
Therefore, the surrogate operator network alleviates the excessive computational burden associated with numerical solvers that are forced to fix a sufficiently small time scale for accurately resolving the fast dynamics during time integration.
Nevertheless, longer forecasts in time still remain a challenge for fast data-driven surrogate networks in general due to the gradual but consistent accumulation of errors.
}

{
For extrapolating in time, direct numerical solvers like MEMPHIS will be the best choice on the basis of accuracy. 
However, the finite-difference-based solver is computationally more expensive when required to generate a trajectory from a new initial condition.
Conversely, the surrogate network trained offline offers faster solutions at inference but suffers in prediction accuracy due to the accumulation of errors. 
To circumvent the predicament associated with either end of the accuracy-speed spectrum, we coupled the numerical solver with the trained surrogate operator network. 
Specifically, we developed a hybrid solver that segues between the direct numerical solver and the trained U-Net conditioned on time during the course of the trajectory. 
From our results, we infer that running the direct numerical solver for longer timesteps helps decreasing the rate of accumulation of error in time.
The surrogate conditioned U-Net by itself is able to restrict the mean relative $L^2-$error of the spatial autocorrelations of the microstructure within 6\% for forecasts up to $t=50$, with a speed-up of 16.5$\times$.
Integrating the surrogate U-Net conditioned on time with the numerical solver MEMPHIS, helps bringing the mean rel. $L^2-$error below 2\%, while retaining a speed-up of 1.3$\times$.
}

{
The hybrid formulation proposed here is generic and can be easily extended to diverse applications.
Based on their performance and accuracy, such hybrid solvers can be utilized for solving data-hungry inverse design problems, \textit{i.e.} for a given observed final state of the materials simulation what are the corresponding input parameters that lead to that state. For instance, prior works \cite{desai2022learning} have relied on iterative methods comprising the genetic algorithm as the inverse model and the MEMPHIS solver as the forward model for 
learning novel physical vapor deposition protocols to design thin films. The bottleneck in this framework is the excessive computational cost associated with executing the forward run several times during the iterative process. Our hybrid formulation would be able to substantially speed up the overall inverse control process by significantly alleviating the associated computational costs.
Another interesting avenue would be to use such hybrid formulation in the context of multi-fidelity modeling frameworks to integrate the numerical and experimental data, where the U-Net hybrid model would serve as the low-fidelity rapid model fused with high-fidelity experimental data.
While we demonstrated the applicability of the hybrid solver on a specific problem, this model can be applied to many other scientific domains such as high-speed flows, climate modeling, reaction-diffusion modeling, and many more.
}

\section{Materials and Methods}
\label{sec: Materials and Methods}

\subsection{Phase-field modeling of physical vapor deposition}
\label{subsec: pvd_theory}
{
The phase-field model used to simulate PVD is described in detail elsewhere~\cite{stewart2020microstructure}.
This model accounts for various mechanisms occurring during the physical vapor deposition and growth of thin films (surface and bulk diffusion, transport of vapor phase) and has been validated experimentally.
In this model, the deposition of thin films is simulated as the evolution of two field variables $c(x,t)$ and $\phi(x,t)$, where $c(x,t)$ represents the concentration of one of the atomic species in the alloy and $\phi (x, t)$ represents the location of the solid-vapor moving interface. The solid phase corresponds to $\phi=1$ and the vapor phase to $\phi=-1$.
The evolution of the composition field (slow dynamics in this study) follows the Cahn-Hilliard equation,
\begin{equation}
    \frac{\partial {c}}{\partial {t}} = \nabla \cdot \left[M(c, \phi) \nabla \frac{\delta {F}}{\delta {c}}\right],
\end{equation}
where $F$ is the free-energy functional defined in Ref.~\cite{stewart2020microstructure} and $M_c(\phi,c)$ is the structurally and compositionally dependent mobility function which captures the relative difference between bulk and surface mobilities such that,
\begin{equation}
    M_c(\phi,c) = M^{\text{Bulk}} + M^{\text{Surf}},
\end{equation}
with
\begin{equation}
    M^{\text{Bulk}} = \frac{1}{4}(2-\phi)(1+\phi)^2\Bigl( h(c)M_A^{\text{Bulk}} 
 + (1-h(c)) M_B^{\text{Bulk}}\Bigr),
\end{equation}
and
\begin{equation}
    M^{\text{Surf}} = \exp{(-(\phi/\sigma^{\text{Surf}})^2) } \Bigl(  h(c)M_A^{\text{Surf}} + (1-h(c))M_B^{\text{Surf}} - M^{\text{Surf}}  \Bigr).
\end{equation}
The function $h(c)$ is a smooth swithching function (see Ref.\cite{stewart2020microstructure} for expression).
}

{
The growth of the solid thin film is described by the Cahn–Hilliard equation with a source term that describes the deposition of new species on the surface of the growing film. As such, the evolution of the vapor-solid interface (fast dynamics in this study) is given by,
\begin{equation}
    \frac{\partial {\phi}}{\partial {t}} = \nabla \cdot [M(\phi) \nabla \frac{\delta {F}}{\delta {\phi}}] + S(\Vec{n}(\phi)),
\end{equation}
where the source term $S$ is defined as:
\begin{equation}
    S(\Vec{n}(\phi)) = \nabla \phi \cdot (\rho \bm{v}),
\end{equation}
with $v$ as the velocity field of the incident vapor flux, $\rho$ as the local vapor phase density field, and $\Vec{n} = \nabla \phi/|\nabla \phi|$ as the interface normal direction..
The transport and evolution of the vapor phase is modeled using a convection-diffusion equation such that,
\begin{equation}
    \frac{\partial \rho}{\partial t} = \nabla \cdot \Bigl(  D_{\rho} \nabla \rho  \Bigr) - \nabla \cdot (\rho \bm{v}) - \nabla \phi \cdot (\rho \bm{v}),
\end{equation}
where $D_{\rho}$ is the mass diffusivity of vapor.
}

\subsection{Dataset}
\label{subsec: dataset}
{
The phase-field model described above has been implemented, validated, and verified in MEMPHIS, a finite-difference-based solver written in modern Fortran, with second-order-central-difference stencils for all spatial derivatives and an explicit midpoint method for time integration.
}

{
We generated 2200 microstructure evolution trajectories by varying several simulation parameters: the deposition rate ($v$) from 0.3 to 1.0 (in simulation units), the angle of deposition from 50$^\circ$ to 90$^\circ$, species mobilites ($M_A, M_B$). The initial phase fraction ($c_A$) was kept constant and set to 50\%.
All simulations in this work are performed on a uniform 2D mesh of $256\times256$ grid points with dimensionless numerical and physical parameters.
Each microstructure evolution trajectory consists of 50 time snapshots equally spaced taken over the entire course of the simulation.
The dataset is thus of shape $2200\times(50, 256, 256)$.
The 2200 trajectories were split into training, validation, and testing datasets in the 80:10:10 ratio.
}

\subsection{U-Net with temporal conditioning}
\label{subsec: unet_arch}

{
The U-Net conditioned on time consists of two networks:
1) a multi-layer perception (MLP) network that learns a suitable basis ($\vec{f}(\Delta t)$) with respect to the scalar parameter - time ($\Delta t$) and
2) a U-Net with element-wise product operation at the residual connections in order to condition the U-Net prediction with respect to the time $\Delta t$.
}
  
{
The MLP used in this study is comprised of 2 hidden layers with 128 neurons each and uses the $\sin$ activation function \cite{leshno1993multilayer}.
The  non-linear basis of time projected by the MLP can be expressed as,
\begin{equation}
    \vec{f}(\Delta t) = w^{MLP}_{2} \sin(w^{MLP}_{1} \sin(w^{MLP}_{0} \Delta t+b^{MLP}_0)+b^{MLP}_1)+b^{MLP}_2,
\end{equation}
where $\{w^{MLP}_i, b^{MLP}_i\}_{i=0}^{2}$ represents the parameters learned during the training of the operator.
} 

{
For the U-Net itself, each convolution block in this network is composed of a two-dimensional convolutional layer \cite{krizhevsky2012imagenet}, a group normalization layer \cite{wu2018group}, and a Gaussian Error Linear Unit (GELU) activation layer \cite{hendrycks2016gaussian}.
Before we formulate the forward pass of the U-Net with temporal conditioning, we define the convolution and group normalization operation for an arbitrary three-dimensional tensor $u$ - with channel size of $C$, width of $W$, and height of $H$. 

{
The two-dimensional convolution can be performed on $u$ using a four-dimensional weight tensor $w^{conv}$ with input channel size of $C$, output channel size of $C'$, width of $W'$, and height of $H'$, in the following manner,
\begin{equation}
    \text{conv}(u)_{k',i,j} = \sum_{k=0}^{C-1} \sum_{m=0}^{W'-1} \sum_{n=0}^{H'-1} u_{k, i+m, j+n} . w^{conv}_{k,k',m,n},
\end{equation}
where $w^{conv}$ is the weight tensor associated with the convolution operation, learned during the course of training the operator network.
}

{
For the group normalization operation on a three-dimensional tensor $u$, we separate $C$ channels to $G$ groups of $\tilde{C}$ channels ($C=G\times\tilde{C}$), compute $G$ means and standard deviations separately as,
\begin{equation}
    \mu_g = \frac{1}{\tilde{C}.W.H}\sum_{\tilde{k}=0}^{\tilde{C}-1} \sum_{m=0}^{W-1} \sum_{n=0}^{H-1} u_{g,\tilde{k},m,n} \quad \text{, $g$=0,1,...,$G$-1},
\end{equation}
\begin{equation}
    \sigma_g^2 = \frac{1}{\tilde{C}.W.H} \sum_{\tilde{k}=0}^{\tilde{C}-1} \sum_{m=0}^{W-1} \sum_{n=0}^{H-1} (u_{g,\tilde{k},m,n} - \mu_g)^2 \quad \text{, $g$=0,1,...,$G$-1}.
\end{equation}
We then normalize $u$ as,
\begin{equation}
    \hat{u}_{g,\tilde{k},i,j} = \frac{u_{g,\tilde{k},i,j} - \mu_g}{\sqrt{\sigma_g^2 + \epsilon}} \quad \text{, $g$=0,1,...,$G$-1} 
\end{equation}
where $\epsilon$ is a small positive constant. 
The output of the group normalization operation is,
\begin{equation}
    \text{GN}(u)_{k,i,j} = \gamma_{k} . \hat{u}_{k,i,j} + \beta_k \quad \text{, $k$=0,1,...,$C$-1}.
\end{equation}
Here, $\gamma_k$ and $\beta_k$ are trainable $C$ dimensional parameters to learn the ideal shift and scale operation.
}

{
The conv($u$) and GN($u$) are linear transformations of $u$.
We introduce non-linearity using GELU activation function, which can be approximately represented as, 
\begin{equation}
    \text{GELU}(u) = 0.5 u (1 + \tanh(\sqrt{\frac{2}{\pi}}  (u + 0.044715 u^3))).
\end{equation}
The convolutional block that non-linearly transforms $u$ can be represented as,
\begin{equation}
    \text{conv\_block}(u) = \text{GELU}(\text{GN}(\text{conv}(u))).
\end{equation}
}

{
The downsampling operation is performed by the two-dimensional max-pooling layer \cite{yamaguchi1990neural}. 
This operation is expressed as,
\begin{equation}
    \text{down}(u)_{k,i,j} = \max_{0\leq m < W'} \max_{0\leq n < H'} u_{k,i.S_w+m, j.S_h+n},
\end{equation}
where $S_w$ and $S_h$ represent strides along width and height.
In order to obtain a scale-down factor of 2 during the downsampling operation $S_w=S_h=W'=H'=2$ during maxpooling.
The up-sampling is performed by the two-dimensional transpose convolutional operation \cite{dumoulin2016guide} in the following manner,
\begin{equation}
    \text{up}(u)_{k,i,j} = \sum_{m=0}^{W'-1} \sum_{n=0}^{H'-1} u_{k,i.S_w+m, j.S_h+n}.w^{tconv}_{m,n},
\end{equation}
where $w^{tconv}_{m,n}$ is a trainable two-dimensional tensor with width $W'$ and height $H'$.
To achieve a scale-up factor of 2, $S_w=S_h=W'=H'=2$.
}

{
Next, we mathematically represent the output $c(t+\Delta t)$ of our U-Net with temporal conditioning ($\mathcal{G}$) that takes $[c(t-lb+1), ..., c(t)]$ with $lb$ channels and $\Delta t$ as the inputs.
The latent variables in the latent space $\mathcal{L}_p$ ($p=1,2,3,4$) before and after temporal conditioning are represented as $\vec{z}^{\mathcal{L}_p}_t$ and $\vec{z}^{\mathcal{L}_p}_{t+\Delta t}$ respectively.
$\vec{z}^{\mathcal{L}_p}_t$ can be expressed as,
\begin{equation}
    \vec{z}^{\mathcal{L}_p}_t = \vec{z}^{\mathcal{L}_p}(t) =
    \begin{cases}
         \text{conv\_block}([c(t-lb+1), ..., c(t)]) & \text{if $p$=1}\\
         \text{conv\_block}(\text{down}(\vec{z}^{\mathcal{L}_{p-1}}_t)) & \text{if $p$=2,3,4}
    \end{cases}
\end{equation}
The temporal conditioning operation that makes the latent variable $\vec{z}$ a function of $\Delta t$ is an element-wise product ($\odot$) as shown below.
\begin{equation}
    \vec{z}^{\mathcal{L}_p}_{t+\Delta t} = \vec{z}^{\mathcal{L}_p}_{t} \odot w^{\mathcal{L}_p}\vec{f}(\Delta t) \\ \implies (z^{\mathcal{L}_p}_{t+\Delta t})_{k,i,j} = (z^{\mathcal{L}_p}_{t})_{k,i,j} . (w^{\mathcal{L}_p}\vec{f}(\Delta t))_k \quad \forall p.
\end{equation}
In the above expression, $w^{\mathcal{L}_p}$ projects the shared non-linear basis, $\vec{f}(\Delta t)$, linearly to number of channels at the $p^{th}$ latent level before conditioning the latent representation.
After conditioning the latent vector on time, we up-sample as follows,
\begin{equation}
    \vec{d}^{\mathcal{L}_p}_{t+\Delta t} =
    \begin{cases}
        \text{up}(\text{conv\_block}(\vec{z}^{\mathcal{L}_{p+1}}_{t+\Delta t})) & \text{if $p$=3}\\
        \text{up}(\text{conv\_block}(\vec{z}^{\mathcal{L}_{p+1}}_{t+\Delta t}\ocircleplus \vec{d}^{\mathcal{L}_{p}}_{t+\Delta t} )) & \text{if $p$=1,2},
    \end{cases}
\end{equation}
where $\ocircleplus$ concatenates the tensors along the dimension of channels.
The output of the model is computed as, 
\begin{equation}
    c(t+\Delta t) = \text{GN}(\text{conv}(\text{conv\_block}(\vec{z}^{\mathcal{L}_1}_{t+\Delta t}\ocircleplus \vec{d}^{\mathcal{L}_{1}}_{t+\Delta t}) ))
\end{equation}
The illustration of the architecture used in this study is provided in \autoref{fig: unet_arch}, part c.
}

{
We implemented and trained the surrogate operator in TensorFlow \cite{abadi2016tensorflow}.
Conditioning of the U-Net with respect to the scalar parameter time is motivated by Gupta \emph{et al.} \cite{gupta2022towards}.
Additionally, our implementation learns the dynamics of evolution by taking advantage of the information contained in the microstructure's composition field at $lb$ previous timesteps as shown in \autoref{eq: surrogate_formulation}.
We demonstrate the significance of incorporating the historical information empirically in the error heatmaps in Appendix \autoref{fig: traj_splitting_heatmaps}, and decided to provide the composition field states at $lb=3$ previous timesteps.
}

\subsection{Training of the network}
\label{subsec: training_details}
{
We used the mini-batch training mode \cite{goodfellow2016deep} with a batch size of 5 for training the U-Net conditioned on time using the Adam optimizer \cite{kingma2014adam} with a learning rate of $10^{-4}$.
The best model was chosen on the basis of the relative mean squared error of the model prediction with respect to the trajectories in the validation dataset.
The model was trained across 8$\times$24GB NVIDIA GeForce RTX 3090 GPUs. 
Since the batch size is much smaller than the total number of samples in the training dataset, we observe that the model converges around the 200 epoch. 
Therefore, the model was trained only for 300 epochs, with a wall time of 14 hours.
}


\subsection{Error metrics}
\label{subsec: error_metrics}
{
The operator networks used in this study were trained to minimize the relative mean squared error (Rel. MSE) between the true and predicted composition fields($c$).
The Rel. MSE can be computed in the following manner,
\begin{equation}
    \text{Rel. MSE}(c) = \sum_n \sum_x \sum_y \sum_t \frac{ (c_{\text{true}}^{(n)}(x,y,t) -  c_{\text{pred}}^{(n)}(x,y,t))^2 }{ (c_{\text{true}}^{(n)}(x,y,t))^2},
\end{equation}
where $n$ corresponds to the $n^{\text{th}}$ sample in the dataset.
}

{
The spatial autocorrelation of the composition field is a better statistical representation of the underlying microstructures as explained in Section~\ref{subsec: autocorrelations} and elsewhere~\cite{herman2020data}.
Therefore, we first compute the spatial autocorrelations of the true and predicted composition fields and then quantify the statistical error on the basis of the relative $L^2$ error between the true and predicted autocorrelation.
The relative $L^2$ error of the spatial autocorrelations can be computed as shown below,
\begin{equation}
    \text{Rel. }L^2(\bm{S}_2^{(A,A)}(\bm{r},t)) = \frac{|| (\bm{S}_2^{(A,A)}(\bm{r},t))_{\text{true}} - (\bm{S}_2^{(A,A)}(\bm{r},t))_{\text{pred}}  ||_2}{|| (\bm{S}_2^{(A,A)}(\bm{r},t))_{\text{true}} ||_2}
\end{equation}
}

\newpage
\section*{Acknowledgements}
%
V.O. acknowledges Alan John Varghese, Dr. Aniruddha Bora, Zhen Zhang, and Dr. Johannes Brandstetter for their insightful suggestions and valuable discussions at various stages of this project.
The phase-field framework is supported by the Center for Integrated Nanotechnologies (CINT), an Office of Science user facility operated for the U.S. Department of Energy. This research was conducted using computational resources and services at the Center for Computation and Visualization, Brown University.
This article has been authored by an employee of National Technology \& Engineering Solutions of Sandia, LLC under Contract No. DE-NA0003525 with the U.S. Department of Energy (DOE).
The employee owns all right, title and interest in and to the article and is solely responsible for its contents. The United States Government retains and the publisher, by accepting the article for publication, acknowledges that the United States Government retains a non-exclusive, paid-up, irrevocable, world-wide license to publish or reproduce the published form of this article or allow others to do so, for United States Government purposes.
The DOE will provide public access to these results of federally sponsored research in accordance with the DOE Public Access Plan \url{https://www.energy.gov/downloads/doe-public-access-plan}.

\noindent \textbf{Funding:} 
R.D. and G.E.K. acknowledge funding under the \textit{Beyond} Fingerprinting Sandia Grand Challenge Laboratory Directed Research and Development (GC LDRD) program. 
Sandia National Laboratories is a multi-mission laboratory managed and operated by National Technology and Engineering Solutions of Sandia, LLC., a wholly owned subsidiary of Honeywell International, Inc., for the U.S.\ Department of Energy National Nuclear Security Administration under contract DE-NA0003525. This paper describes objective technical results and analysis. Any subjective views or opinions that might be expressed in the paper do not necessarily represent the views of the U.S.\ Department of Energy or the United States Government.   
K.S. and G.E.K. were partially funded by the Air Force Office of Science and Research (AFOSR) under OSD/AFOSR MURI Grant FA9550-20-1-0358.

\noindent \textbf{Author Contributions:} R.D. and G.E.K. designed the study and supervised the project. V.O. developed the autoencoder-based models and U-Net with temporal conditioning. V.O. and K.S. parallelized the code across GPUs. V.O., K.S., and R.D. coupled the surrogate operator with MEMPHIS. V.O., K.S., and S.D. analyzed the results. V.O., K.S., R.D., and G.E.K. wrote the manuscript.

\noindent \textbf{Competing Interests:} The authors declare no competing interests.

\noindent \textbf{Data and Materials Availability:} The data that support the findings of this study are available from the corresponding authors upon reasonable request. The codes of the autoencoder and U-Net-based operators used in this study were developed in Python3 and will be made available at this \href{https://github.com/vivek-brown/UNet-with-temporal-conditioning} {git repository} after acceptance.

\newpage
\bibliographystyle{unsrt}  
\bibliography{references}  
\newpage

\appendix
\counterwithin{figure}{section}
\setcounter{figure}{0}

\section{Analyzing the latent dimensionality for spatial reconstruction}
{
In this supplementary note, we investigate the effect of the dimensionality of the latent space of the encoder for the autoencoder-based models on the reconstruction of the microstructures. 
We train seven different autoencoders with latent dimensions of ld = 64, 100, 196, 256, 400, 625, and 900. 
The autoencoder consists of an encoder that learns to compress the microstructure composition field ($c(x,y)$) to a lower dimensional embedding ($\bm{\tilde{c}}$) and a decoder that learns to reconstruct the composition field from the embedding.
The encoder can be interpreted as a nonlinear dimensionality reduction algorithm. 
Note that we are not trying to learn the evolution dynamics in this experiment, but rather attempting to identify a sufficiently large latent space that can reconstruct the microstructure from its encoded state with an acceptable level of accuracy. 
}
\begin{figure}[H]
  \centering
  \includegraphics[width=0.99\textwidth]{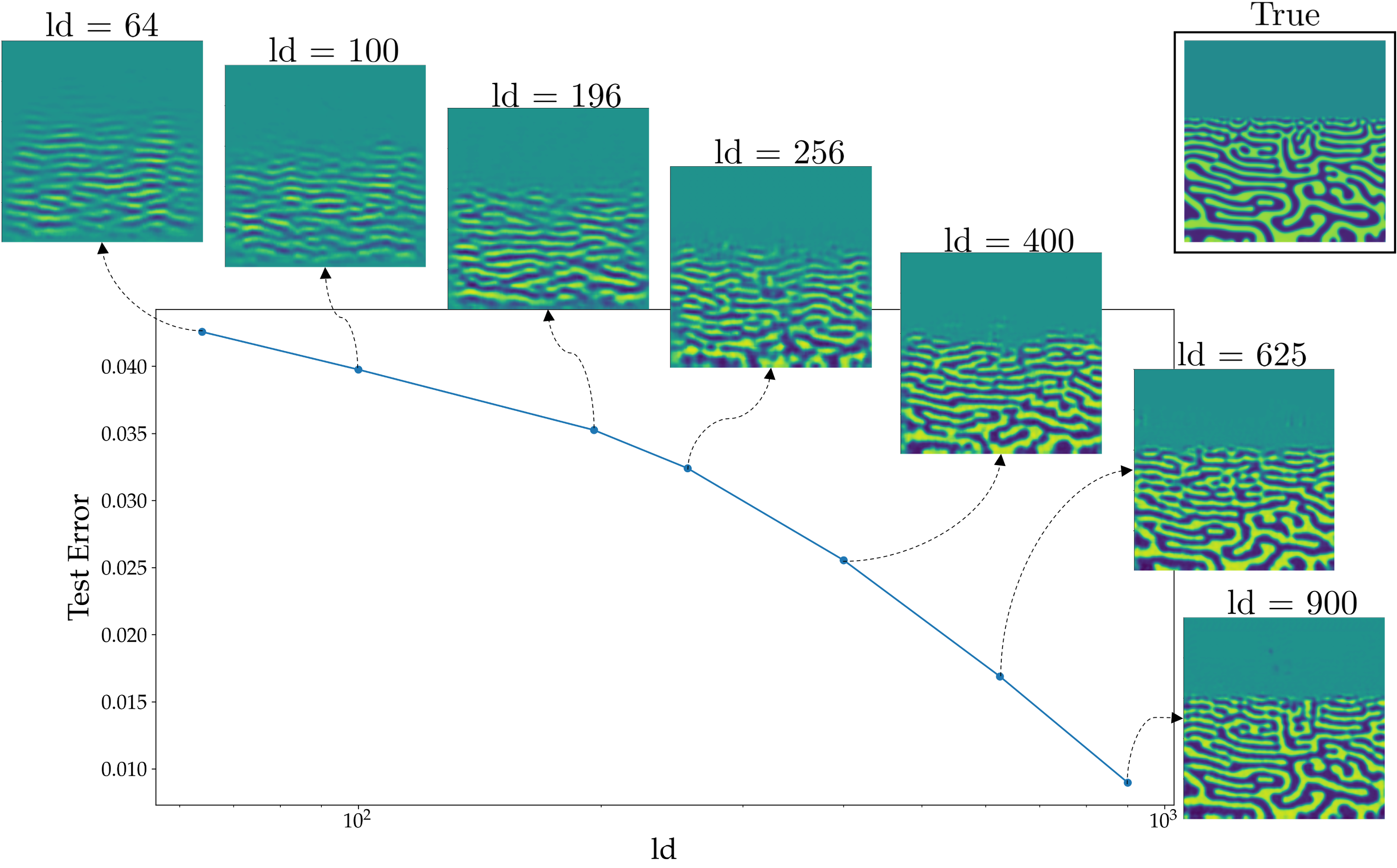}
  \caption{\textbf{Effect of the latent dimension on the reconstruction error.} We investigate the variation in the spatial reconstruction errors in the autoencoder against the dimensionality (ld) of the bottleneck layer. As expected, we observe a decreasing trend.}
  \label{fig: ld_experiment}
\end{figure}
{
From \autoref{fig: ld_experiment}, we observe that the test MSE decreases and the quality of the composition fields reconstructed by the autoencoder improves as the dimensionality of the latent space increases. 
The presence of finer, low-wavelength features inhibits the autoencoder from accurately reconstructing the microstructure, especially for lower dimensionality of the latent space.
Subsequently, in the case of the baseline autoencoder-DeepONet model, the latent DeepONet model is required to learn the evolution dynamics of the microstructure embedding in an ld dimensional latent space. 
In other words, for a larger $ld$, the model's accuracy to reconstruct space increases, but its ability to learn evolution dynamics depreciates. 
This motivated us to investigate the ways of learning latent dynamics for larger values of ld. 
}

\newpage
\section{Visualizing the evolution dynamics of the encoded embeddings}
{
We visualize and analyze the evolution dynamics of the microstructure embeddings ($\bm{\tilde{c}}(t)$) in the latent space learned by the encoder. 
The latent space considered in this study has a dimensionality of $ld=$3137 ($\bm{\tilde{c}}(t) \in \mathbb{R}^{ld=3137}$).
We decided to move ahead with a large $ld$ to ensure accurate reconstruction of microstructure from its embedding, as discussed in the previous note S1.
We perform Singular Value Decomposition (SVD) on $\bm{\tilde{c}}$ using the truncated SVD algorithm \cite{halko2011finding} such that,  
}
\begin{equation}
    \bm{\tilde{c}}(t) = \Bigl( \bm{U\Sigma} \Bigr)(t)\text{  }V,
\end{equation}
{
where $V$ represents $ld-1$ proper orthogonal decomposition (POD) basis/modes that can span all the $\bm{\tilde{c}}(t)$ encountered in the train dataset.
$\Bigl( \bm{U\Sigma} \Bigr)(t) \in \mathbb{R}^{ld-1}$ represents the vector of coefficients at time $t$, such that, a linear combination of the set of POD modes in $V$ with respect to the coefficients $\Bigl( \bm{U\Sigma} \Bigr)(t)$, approximates $\bm{\tilde{c}}(t)$ in the least square sense. 
The latent dynamics can be interpreted to be present in the linear projection of $\bm{\tilde{c}}(t)$ with respect to $V^T$.
}
\begin{figure}[H]
  \centering
  \includegraphics[width=0.99\textwidth]{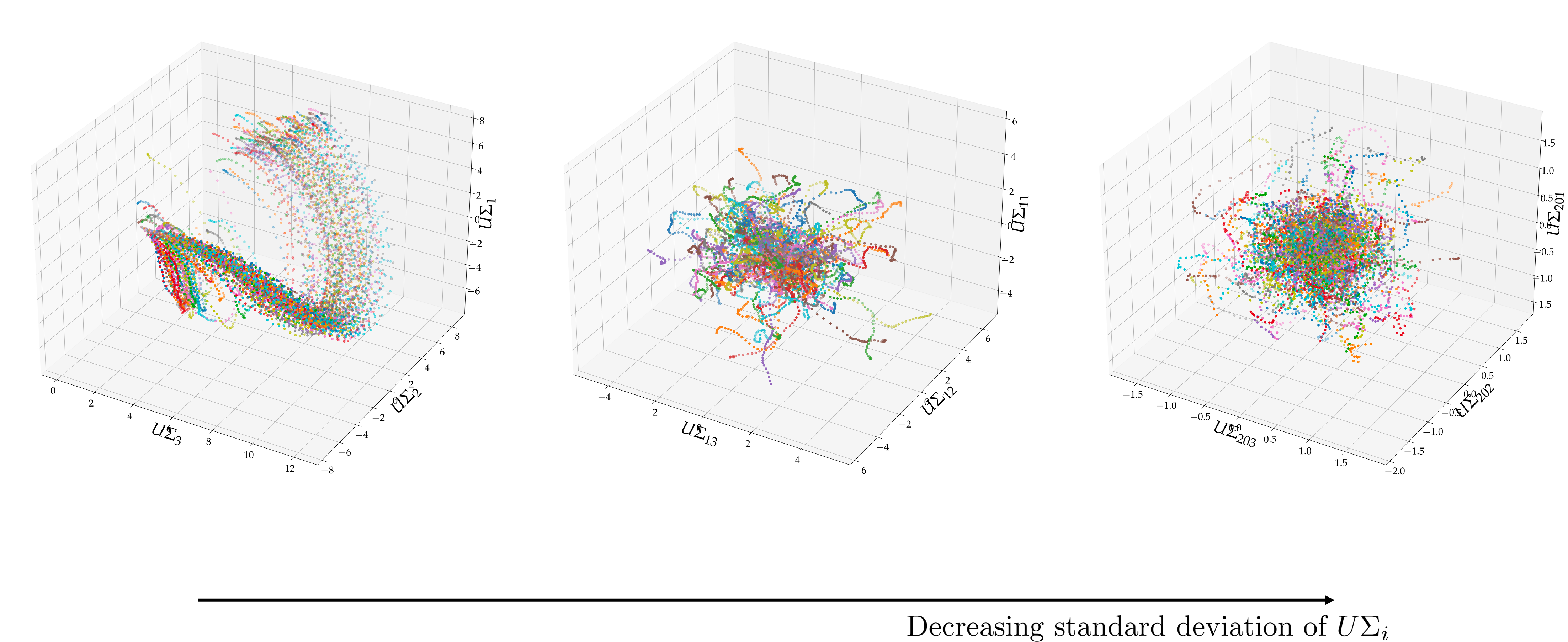}
  \caption{\textbf{Visualizing the evolution dynamics of the encoded microstructure embeddings.} From visual inspection, we observe that the dynamics contained in the higher energy modes evolve smoothly compared to the lower energy modes. The figure motivated us to partition and scale the latent dynamics based on the POD modes.}
  \label{fig: hierarchical_latent_dynamics}
\end{figure}
{
In \autoref{fig: hierarchical_latent_dynamics}, we present scatter plots ($U\Sigma_i$, $U\Sigma_{i+1}$, $U\Sigma_{i+2}$) for $i=$1, 11 and 201 as a way to visualize the dynamic evolution of the embedded microstructure.
We observe that the standard deviation of the $i^{th}$ component of $\bm{U\Sigma}$ decreases with $i$.
This trend suggests that performing the SVD splits the original encoded trajectory of $\bm{\tilde{c}}(t)$ into multiple sub-trajectories $U\Sigma_i(t)$ at different scales. 
This visualization and analysis motivated us to hierarchically fine-tune multiple DeepONet models, to different scales at sets of $i$'s, and formulate the hierarchical autoencoder-DeepONet model. 
The proposed hierarchical model architecture is presented in \autoref{fig: unet_arch} b). 
}

\newpage
\section{Analyzing the partitioning of the latent space}
{
As discussed in the previous subsection, the hierarchical autoencoder-DeepONet fine tunes multiple DeepONet models to different hierarchical partitions of the eigen coefficients $\Bigl(\bm{U \Sigma}\Bigr)(t) \in \mathbb{R}^{3136\times1}$ of the encoded embedding $\bm{\tilde{c}}(t)$ in the latent space of the autoencoder.
Here, we analyze the performance of the model when increasing the number of equal-sized latent partitions with $n = $ 1, 2, 4, 8, 16.
We investigated the quality of model prediction by visualizing the energy of the trajectories in latent space represented by the eigenvalues of the true and predicted latent microstructure embeddings ($\bm{\tilde{c}}(t)$).
}
\begin{figure}[H]
  \centering
  \includegraphics[width=0.9\textwidth]{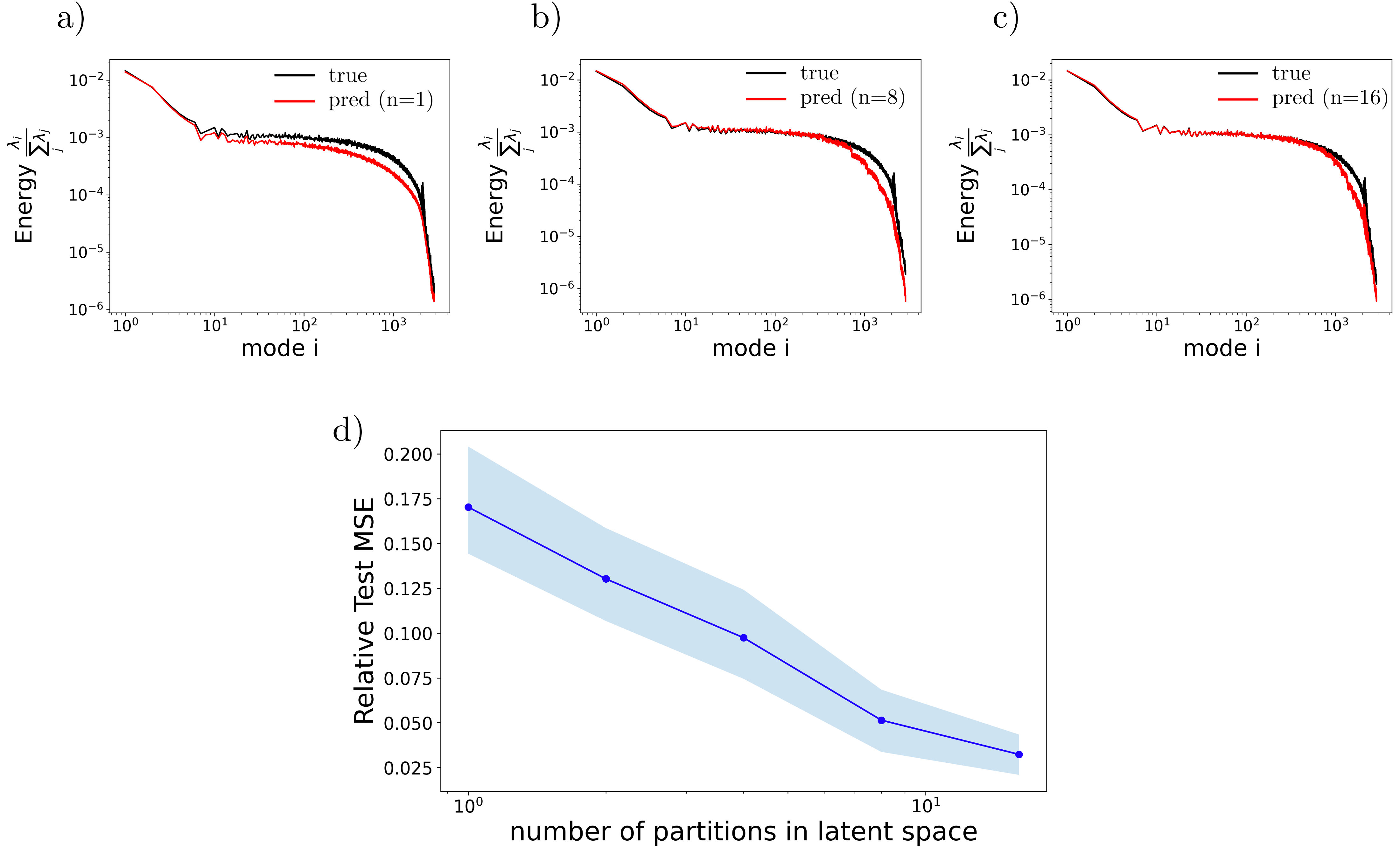}
  \caption{\textbf{Effect of partitioning of the latent space.} Panels a)-c) represent the comparison between the true and predicted energy spectrum of the latent trajectories for different number of latent partitions $n=1,8,16$, respectively. Panel d) provides a statistical representation of the relative test MSE against the number of latent partitions. }
  \label{fig: latent_partition_experiment}
\end{figure}

\begin{table}[H]
    \caption{\textbf{Partitioning of the latent space} }
    \centering
        \begin{tabular}{l||cc}
             & Mean      & Std        \\
        \hline
        \hline
        n=1  & 1.703e-01 & 4.115e-02  \\
        n=2  & 1.303e-01 & 3.498e-02  \\
        n=4  & 9.742e-02 & 3.102e-02  \\
        n=8  & 5.143e-02 & 2.076e-02  \\
        n=16 & 3.232e-02 & 1.351e-02 
        \end{tabular}
    \label{table: latent_partition_experiment}
\end{table}
{
From \autoref{fig: latent_partition_experiment} a) to c), we observe that increasing the number of latent partitions brings the eigenvalues of the predicted embedding closer to that of true embeddings, suggesting more accurate model predictions.
The effect is prominent in the lower eigenmodes that carry more energy. 
The observation is intuitive because \autoref{fig: hierarchical_latent_dynamics} suggests a decay in energy across the modes and portrays smoother and easy-to-learn dynamics for lower modes.
In \autoref{fig: latent_partition_experiment} d) and \autoref{table: latent_partition_experiment}, we investigate the variation in the relative MSE of $\bm{\tilde{c}}(t)$ with respect to all the samples in the test dataset across all time steps, on increasing the number of latent partitions.
Both the mean and standard deviation of the relative test MSE decrease with the number of latent partitions, implying more accurate and confident predictions.
Note that during this experiment we only look into equal-sized latent partitions. 
There could be better ways of partitioning the latent space, and this could be a possible direction for future research.
}

\newpage
\section{Trajectory splitting for supervised learning }
{
Our PVD exemplar is a two-dimensional dynamical system.
In a data-driven learning setting, the dataset consists of two-dimensional trajectories evolving from different initial conditions, but governed by the same PDEs.
The PDEs that govern the PVD process and the numerical methods used for generating the simulation data are provided in Section~\ref{subsec: pvd_theory}.
In the context of data-driven deep learning of two-dimensional dynamical systems, each sample trajectory can be further split into sub-trajectories because the PDEs remains unchanged.
Therefore, it is reasonable to train the surrogate operator network ($\mathcal{G}$) to learn the mapping from a look-back time window of size $lb$ time steps to a look-forward window of size $lf$ time steps, as shown below,
\begin{equation}
    [c_{t-lb+1}, ..., c_t] \rightarrow [c_{t+1}, ..., c_{t+lf}]. 
\end{equation}
\begin{equation}
    c(t+\Delta t) \approx \mathcal{G}([c_{t-lb+1}, ..., c_t])(\Delta t), \text{ where $\Delta t \in \{1,2,...,lf\}$}
\end{equation}
This will increase the number of samples available for training the surrogate network.
}

{
To investigate the optimal choice of $lb$ and $lf$, we perform an analysis by varying both the look-back and look-forward time windows with $lb$ = 1,3,5,7,9 and $lf$ = 3,5,7,9.
From \autoref{fig: traj_splitting_heatmaps} a) we observe that for a given $lf$, the error is similar for $lb$ = 3, 5, 7 and 9.
However, there is a sudden increase in error when $lb$ = 1.
\autoref{fig: traj_splitting_heatmaps} b), c), d) and e) represents the mean relative MSE on the test dataset at $t_i+3$, $t_i+5$, $t_i+7$ and $t_i+9$ respectively.
Intuitively, these results suggest an increasing trend in error while forecasting further into the future.
The higher error for $lb=1$ suggests that the physical vapor deposition is behaving like a non-Markovian process inflicted by the larger step-size $\Delta t$ used for training the surrogate operator.
Note that the error remains unchanged for $lb \geq 3$, justifying that $lb=3$ should be the number of previous states of the composition field incorporated in training.
}
\begin{figure}[H]
  \centering
  \includegraphics[width=0.65\textwidth]{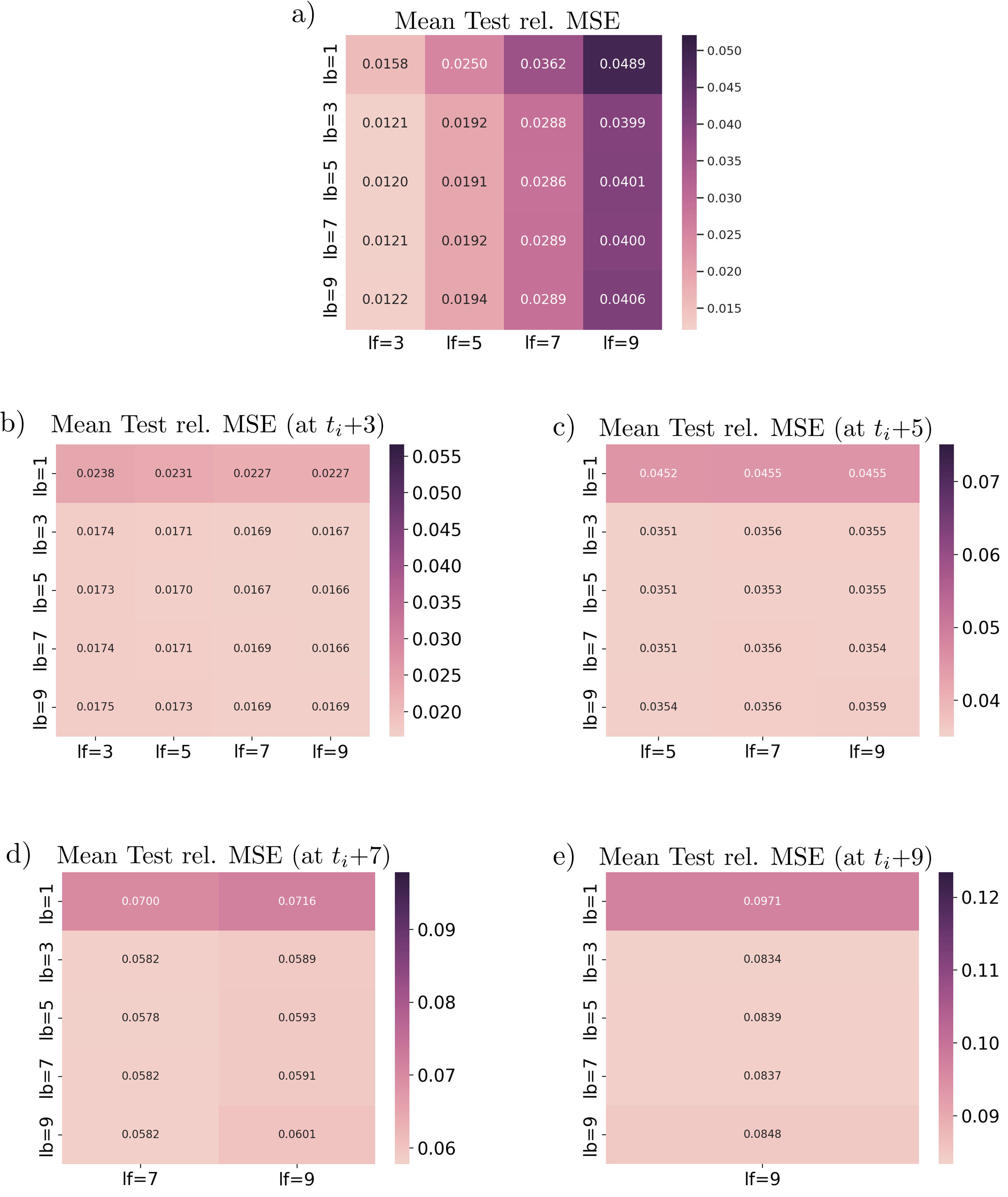}
  \caption{\textbf{Error heatmaps.} Error heatmaps with respect to the relative MSE corresponding to the different look-back ($lb$) and look-forward ($lf$) windows considered in this study.}
  \label{fig: traj_splitting_heatmaps}
\end{figure}

\end{document}